\documentclass[cameraready]{Interspeech}
\usepackage{booktabs}
\usepackage{multirow}
\usepackage{graphicx}
\usepackage[table]{xcolor}
\usepackage{amssymb} 
\usepackage{pifont}  
\usepackage{subcaption}

\title{Voice of  India: A Large-Scale Benchmark for Real-World Speech Recognition in India}

\author[affiliation={1}, orcid=0009-0004-7409-0034, equalcontribution]{Kaushal}{Bhogale}
\author[affiliation={2}, orcid=0009-0002-4923-1486, equalcontribution]{Manas}{Dhir}
\author[affiliation={2}, equalcontribution]{Amritansh}{Walecha}
\author[affiliation={2}]{Manmeet}{Kaur}
\author[affiliation={2}]{Vanshika}{Chhabra}
\author[affiliation={2}]{Aaditya}{Pareek}
\author[affiliation={2}]{Hanuman}{Sidh}
\author{Mahima}{Manik}
\author[affiliation={2}]{Sagar}{Jain}
\author[affiliation={2}]{Bhaskar}{Singh}
\author[affiliation={2}]{Utkarsh}{Singh}
\author[affiliation={1}]{Tahir}{Javed}
\author[affiliation={2}]{Shobhit}{Banga}
\author[affiliation={1}]{Mitesh M.}{Khapra}

\address{
    $^1$ Indian Institute of Technology, Madras, India \quad
    $^2$ Josh Talks, India
}
\email{shobhit@joshtalks.com}


\newcommand{\sarvamaudio}{{\textsc{Sarvam Audio}}}
\newcommand{\sarvamsaarika}{{\textsc{Saarika 2.5}}}

\newcommand{\geminipro}{{\textsc{Gemini 3 Pro}}}
\newcommand{\geminiflash}{{\textsc{Gemini 3 Flash}}}

\newcommand{\gptfouro}{{\textsc{GPT-4o Transcribe}}}
\newcommand{\gptfouromini}{{\textsc{GPT-4o Mini Transcribe}}}

\newcommand{\amazontranscribe}{{\textsc{Amazon Transcribe}}}
\newcommand{\deepgramnova}{{\textsc{Deepgram Nova 3}}}
\newcommand{\assemblyai}{{\textsc{AssemblyAI Universal}}}
\newcommand{\elevenlabscribe}{{\textsc{ElevenLabs Scribe v2}}}
\newcommand{\microsoftstt}{{\textsc{Microsoft Speech-to-Text}}}

\newcommand{\indicconformer}{{\textsc{Indic Conformer}}}

\newcommand{\omniaasroneb}{{\textsc{OmniASR LLM 1B}}}
\newcommand{\omniaasrsevenb}{{\textsc{OmniASR LLM 7B}}}

\usepackage{comment}

\begin{document}

\maketitle


\begin{abstract}
    \sloppy
    \sloppy
Existing Indic ASR benchmarks often use scripted, clean speech and leaderboard driven evaluation that encourages dataset specific overfitting. In addition, strict single reference WER penalizes natural spelling variation in Indian languages, including non standardized spellings of code-mixed English origin words. To address these limitations, we introduce Voice of India, a closed source benchmark built from unscripted telephonic conversations covering 15 major Indian languages across 139 regional clusters. The dataset contains 306230 utterances, totaling 536 hours of speech from 36691 speakers with transcripts accounting for spelling variations. We also analyze performance geographically at the district level, revealing disparities. Finally, we provide detailed analysis across factors such as audio quality, speaking rate, gender, and device type, highlighting where current ASR systems struggle and offering insights for improving real world Indic ASR systems.
    \fussy
    \fussy
\end{abstract}

\section{Introduction}

\noindent Recent progress in Indic Automatic Speech Recognition (ASR) has been driven by shared tasks and large scale benchmarks such as MUCS \cite{diwan2021}, IndicSUPERB \cite{javed2023indicsuperb}, Vistaar \cite{bhogale2023}, and datasets like IndicVoices \cite{javed2024indicvoices}, which have expanded coverage across languages, accents, orthographies, and code switching. However, improvements on benchmark leaderboards often fail to translate to robust real world performance. Existing benchmarks remain cleaner and more scripted than production audio \cite{likhomanenko2020rethinking}, and typically report only a single aggregate WER per language, masking large performance differences across regions and dialects. Their public leaderboard structure further encourages dataset specific optimization, rewarding models that exploit benchmark artifacts rather than generalize to real conversational speech. This issue is amplified by the reliance on a single reference transcript and strict WER scoring, which penalizes legitimate orthographic variation, including spelling differences and non standardized native script renderings of English origin words in code mixed spontaneous speech. 

\begin{figure}[t]
    \centering
    \includegraphics[width=0.7\linewidth]{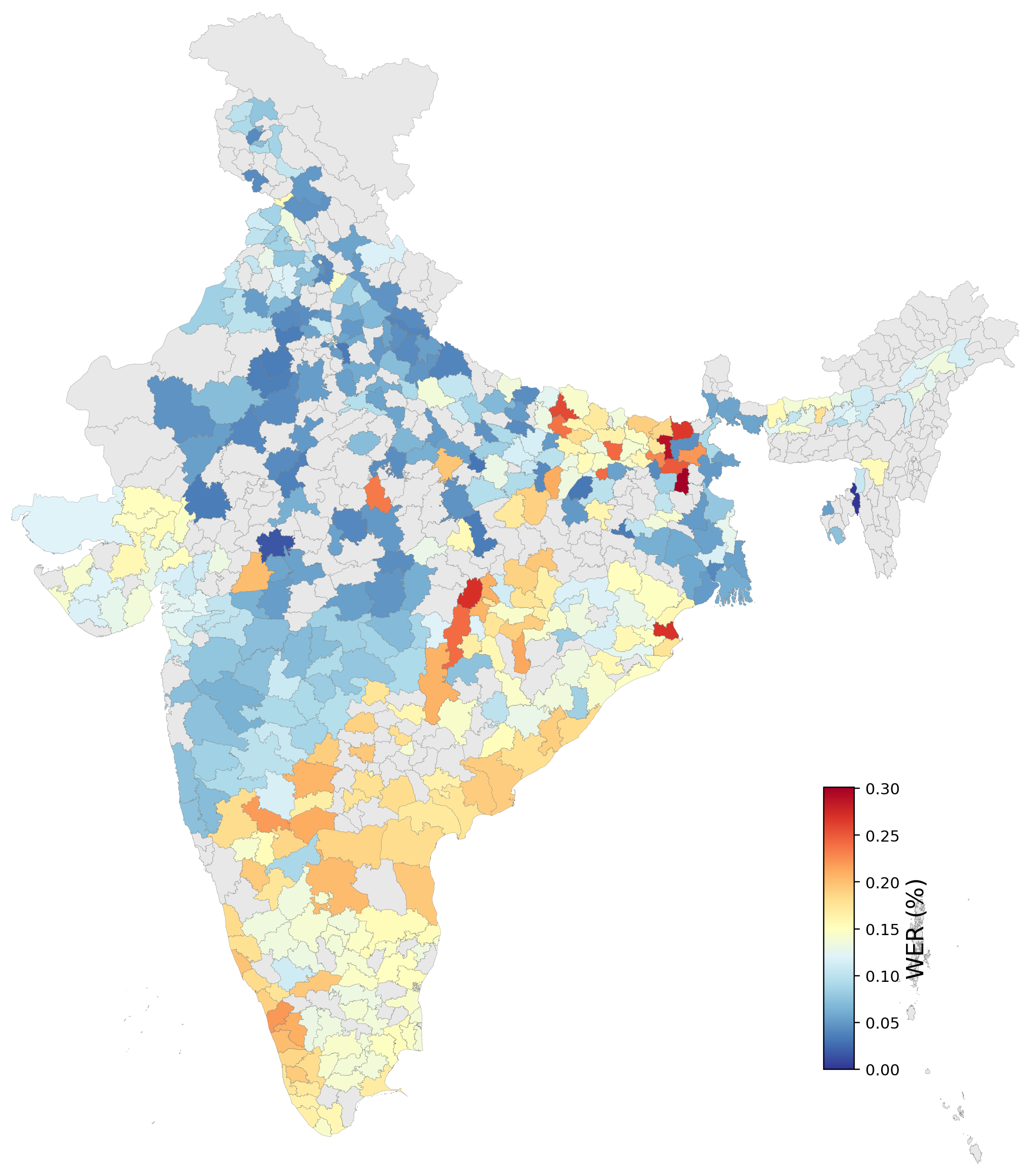}
    \caption{The WER map of India: Average Word Error Rate (WER) for ASR models for districts of India}
    \label{map}
\end{figure}

To address these gaps, we introduce Voice of India, a closed source evaluation benchmark built from unscripted, long form telephonic conversations that reflect how speech naturally occurs in everyday Indian interactions. The benchmark is designed to evaluate ASR systems on spontaneous speech rather than scripted prompts, emphasizing semantic faithfulness over rigid string matching. To avoid penalizing legitimate orthographic variation, the dataset includes multiple valid transcripts that capture natural spelling differences and alternative renderings commonly found in spontaneous and code mixed speech. A central goal of the benchmark is to expose geographic disparities in performance. Accordingly, we analyze results at a regional level: Figure \ref{map} visualizes district level WER, averaged across languages within each region and across four models that support all fifteen languages, providing a nationwide view of the error rates experienced by users.

The dataset is constructed using a population proportional cluster sampling strategy to ensure balanced geographic representation. Specifically, we consider 139 regional clusters covering nearly the entire country, and sample utterances from each cluster in proportion to its population. The resulting corpus spans 15 major Indian languages and contains 306,230 utterances totaling 536 hours of speech from 36,691 speakers. Beyond aggregate metrics, we perform detailed analyses across multiple factors including audio quality, speaking rate, segment length, geographic region, gender, recording device, and age group. This fine grained evaluation is also to identify the specific conditions and regions where existing ASR models lack robustness, thereby informing future research and model development. We release an open subset via a public submission pipeline\footnote{\url{https://github.com/JoshTalks/voice-of-india}}; private test-set evaluation is available on request through our project page - \footnote{\url{https://voiceofindia.ai/}}.

\section{Related Work}

\noindent\textbf{Benchmarks for Indian Language ASR.}
Early efforts to evaluate ASR for Indian languages include the Interspeech 2018 Low Resource ASR Challenge \cite{srivastava2018}, multilingual speech corpora released through OpenSLR \cite{he2020,butryna2020}, and the MUCS 2021 shared task \cite{diwan2021}.More recent benchmarks include IndicSUPERB \cite{javed2023indicsuperb}, Vistaar \cite{bhogale2023}, accent focused datasets such as Svarah \cite{javed2023svarah} and Lahaja \cite{lahaja2024}, and large scale demographically diverse datasets such as IndicVoices \cite{javed2024indicvoices}, covering all 22 scheduled Indian languages.

\noindent\textbf{Large-Scale Speech Data Collection.}
Mozilla Common Voice \cite{ardila2020} pioneered crowdsourced multilingual speech collection, while Google through their OpenSLR effort released multiple datasets for under-resourced languages \cite{butryna2020,kjartansson2018}. FLEURS \cite{conneau2022} expanded multilingual coverage with a standardized evaluation dataset, and the WAXAL initiative \cite{waxal2025} collected speech resources for several Sub-Saharan African languages. Across these efforts, challenges include community mobilization, scalable quality control, and standardized data collection protocols.

\noindent\textbf{ASR Evaluation Beyond WER.}
Multi reference alignment methods \cite{arabic2015,arabic2019,japanese,style_agnostic} reduce spurious penalties from spelling variation but require expensive multiple transcriptions. Rule based substitution frameworks such as SCLITE \cite{sclite} support explicit variant mappings, yet rely on exhaustive enumeration that is impractical for languages with extensive spelling variation and code mixing. Alternatives such as WERd \cite{werd}, normalization based evaluation \cite{whisper,malayalam_normalization}, and phoneme based metrics \cite{power,snwer} address some limitations but depend on incomplete external resources or normalization schemes \cite{malayalam_complex,malayalam_cer}.

\begin{table}[t]
\centering
\caption{Overall statistics of the Voice of India Benchmark.}
\label{tab:stats}
\resizebox{\columnwidth}{!}{%
\begin{tabular}{lrrrr@{\hskip 1.5em}|lrrrr}
\toprule
\textbf{Lang} & \textbf{Hrs} & \textbf{Dist.} & \textbf{Utt.} & \textbf{Spks} &
\textbf{Lang} & \textbf{Hrs} & \textbf{Dist.} & \textbf{Utt.} & \textbf{Spks} \\
\midrule
hne  & 20.8 & 67  & 11.5K & 349  & ml  & 45.2 & 43  & 20.9K & 1090 \\
mr  & 48.6 & 34  & 24.4K & 4501 & pa  & 32.5 & 58  & 17.9K & 1030 \\
bn  & 50.3 & 54  & 24.2K & 6014 & te  & 51.7 & 43  & 39.2K & 7833 \\
ta  & 53.4 & 37  & 32.6K & 4434 & as  & 23.4 & 29  & 13.8K & 390  \\
hi  & 46.9 & 306 & 22.0K & 4270 & or  & 22.4 & 40  & 15.9K & 390  \\
gu  & 44.8 & 31  & 29.1K & 4634 & mai & 22.2 & 36  & 14.1K & 658  \\
bho  & 23.0 & 99  & 12.7K & 390  & ur  & 26.2 & 119 & 14.7K & 388 \\
kn  & 24.9 & 36  & 12.6K & 320  & \textbf{Total} & 536.1 & 675 & 306.2K & 36.6K \\
\bottomrule
\end{tabular}%
}
\end{table}

\begin{figure*}[!t]
\centering

\begin{subfigure}[c]{0.73\textwidth}
\centering
\scriptsize
\setlength{\tabcolsep}{2pt}

\caption{Results of open source models and public APIs on the Voice of India Benchmark}
\begin{tabular}{lccccccccccccccc}
\toprule
\multicolumn{1}{l}{\textbf{Model}} & \multicolumn{1}{c}{\textbf{as}} & \multicolumn{1}{c}{\textbf{bn}} & \multicolumn{1}{c}{\textbf{bho}} & \multicolumn{1}{c}{\textbf{gu}} & \multicolumn{1}{c}{\textbf{hi}} & \multicolumn{1}{c}{\textbf{hne}} & \multicolumn{1}{c}{\textbf{ka}} & \multicolumn{1}{c}{\textbf{mai}} & \multicolumn{1}{c}{\textbf{ml}} & \multicolumn{1}{c}{\textbf{mr}} & \multicolumn{1}{c}{\textbf{or}} & \multicolumn{1}{c}{\textbf{pa}} & \multicolumn{1}{c}{\textbf{ta}} & \multicolumn{1}{c}{\textbf{te}} & \multicolumn{1}{c}{\textbf{ur}} \\
\midrule
\elevenlabscribe                & \cellcolor[HTML]{FFFEFE}15.6    & \cellcolor[HTML]{C7E8D8}10      & \cellcolor[HTML]{FDF2F1}23.5     & \cellcolor[HTML]{FEF6F5}21.2    & \cellcolor[HTML]{ADDEC6}7.7     & \cellcolor[HTML]{FEF7F7}20.3     & \cellcolor[HTML]{FEF9F8}19.2    & \multicolumn{1}{l}{-}            & \cellcolor[HTML]{FDF3F2}23      & \cellcolor[HTML]{E8F5EF}12.9    & \cellcolor[HTML]{FEF7F6}20.7    & \cellcolor[HTML]{FFFEFE}15.6    & \cellcolor[HTML]{FEF7F6}20.4    & \cellcolor[HTML]{FDF3F2}23      & \cellcolor[HTML]{FCEFEE}25.5    \\
\amazontranscribe                           & \multicolumn{1}{l}{-}           & \cellcolor[HTML]{BDE4D1}9.1     & \cellcolor[HTML]{F9DFDD}36       & \cellcolor[HTML]{FFFBFB}17.7    & \cellcolor[HTML]{A3DABF}6.8     & \cellcolor[HTML]{FAE4E2}32.9     & \cellcolor[HTML]{FEFAF9}18.6    & \multicolumn{1}{l}{-}            & \cellcolor[HTML]{FCEBEA}28.2    & \cellcolor[HTML]{D6EEE2}11.3    & \cellcolor[HTML]{FFFBFB}17.9    & \cellcolor[HTML]{FFFEFE}15.9    & \cellcolor[HTML]{FEF9F8}19.3    & \cellcolor[HTML]{FEF8F8}19.7    & \multicolumn{1}{l}{-}           \\
\assemblyai            & \cellcolor[HTML]{E67C73}104.8   & \cellcolor[HTML]{E67C73}103.8   & \cellcolor[HTML]{F6CFCC}46.1     & \cellcolor[HTML]{E67C73}101.8   & \cellcolor[HTML]{FEF9F8}19.3    & \cellcolor[HTML]{F7D3D0}43.6     & \cellcolor[HTML]{EA8D86}89      & \multicolumn{1}{l}{-}            & \cellcolor[HTML]{E67C73}107.5   & \cellcolor[HTML]{EA9088}87.6    & \multicolumn{1}{l}{-}           & \cellcolor[HTML]{E67C73}101     & \cellcolor[HTML]{F3BEBA}57.4    & \cellcolor[HTML]{E67C73}105     & \cellcolor[HTML]{FBE5E4}31.9    \\
\deepgramnova                    & \multicolumn{1}{l}{-}           & \cellcolor[HTML]{FBEAE8}28.9    & \cellcolor[HTML]{F6D0CD}45.8     & \multicolumn{1}{l}{-}           & \cellcolor[HTML]{E9F6F0}13      & \cellcolor[HTML]{F7D5D2}42.4     & \cellcolor[HTML]{F4C4C0}53.7    & \multicolumn{1}{l}{-}            & \multicolumn{1}{l}{-}           & \cellcolor[HTML]{F7D3D0}43.7    & \multicolumn{1}{l}{-}           & \multicolumn{1}{l}{-}           & \cellcolor[HTML]{F0AEA8}67.8    & \cellcolor[HTML]{F7D4D1}43.1    & \multicolumn{1}{l}{-}           \\
\geminipro                       & \cellcolor[HTML]{FEF7F7}20.1    & \cellcolor[HTML]{B6E1CC}8.5     & \cellcolor[HTML]{FEFAFA}18.4     & \cellcolor[HTML]{FFFEFE}15.8    & \cellcolor[HTML]{9AD6B9}6       & \cellcolor[HTML]{FFFCFC}17.2     & \cellcolor[HTML]{FEF8F7}19.9    & \cellcolor[HTML]{FCEFEE}25.6     & \cellcolor[HTML]{FEF5F4}21.7    & \cellcolor[HTML]{CFEBDE}10.7    & \cellcolor[HTML]{FEF6F6}20.9    & \cellcolor[HTML]{F9FCFB}14.4    & \cellcolor[HTML]{FFFEFE}15.7    & \cellcolor[HTML]{FDF5F4}21.9    & \cellcolor[HTML]{BDE4D1}9.1     \\
\geminiflash                     & \cellcolor[HTML]{FCEDEC}26.9    & \cellcolor[HTML]{E5F4EC}12.6    & \cellcolor[HTML]{FDF4F3}22.6     & \cellcolor[HTML]{FDF4F3}22.5    & \cellcolor[HTML]{B4E0CB}8.3     & \cellcolor[HTML]{FDF2F1}23.9     & \cellcolor[HTML]{FDF4F3}22.2    & \cellcolor[HTML]{FBE7E5}30.8     & \cellcolor[HTML]{FCEDEB}27.1    & \cellcolor[HTML]{FFFEFE}16      & \cellcolor[HTML]{FCEEED}26.1    & \cellcolor[HTML]{FEF9F8}19.4    & \cellcolor[HTML]{FEF8F7}19.9    & \cellcolor[HTML]{FCEBEA}27.9    & \cellcolor[HTML]{DDF1E7}11.9    \\
\gptfouro                  & \cellcolor[HTML]{E8857C}94.7    & \cellcolor[HTML]{F7D1CE}44.9    & \cellcolor[HTML]{F5CBC7}49       & \cellcolor[HTML]{E77F76}98.2    & \cellcolor[HTML]{FAE2E0}33.9    & \cellcolor[HTML]{F7D1CE}45.2     & \cellcolor[HTML]{EB958D}84.2    & \cellcolor[HTML]{F2B9B5}60.4     & \cellcolor[HTML]{E78178}97      & \cellcolor[HTML]{F4C1BD}55.6    & \cellcolor[HTML]{EFA7A1}72.5    & \cellcolor[HTML]{EFABA5}70.1    & \cellcolor[HTML]{F1B4AE}64.2    & \cellcolor[HTML]{F0ACA6}69.3    & \cellcolor[HTML]{F9E0DE}35.4    \\
\gptfouromini             & \cellcolor[HTML]{F9DDDA}37.6    & \cellcolor[HTML]{FEF6F5}21.1    & \cellcolor[HTML]{F5CBC7}49.1     & \cellcolor[HTML]{E67C73}295.9   & \cellcolor[HTML]{FEF8F8}19.6    & \cellcolor[HTML]{F7D2CF}44.6     & \cellcolor[HTML]{E78078}97.5    & \cellcolor[HTML]{F6D0CD}45.6     & \cellcolor[HTML]{E67C73}167.8   & \cellcolor[HTML]{FBE7E6}30.7    & \cellcolor[HTML]{F8D6D3}42.1    & \cellcolor[HTML]{F9DCDA}37.9    & \cellcolor[HTML]{F5C7C3}51.9    & \cellcolor[HTML]{EC9992}81.2    & \cellcolor[HTML]{F5C6C2}52      \\
\indicconformer                     & \cellcolor[HTML]{F8FCFA}14.3    & \cellcolor[HTML]{CFEBDE}10.7    & \cellcolor[HTML]{F9E0DE}35.4     & \cellcolor[HTML]{FFFBFA}18      & \cellcolor[HTML]{B3E0CA}8.2     & \cellcolor[HTML]{FBE6E4}31.6     & \cellcolor[HTML]{FEF5F5}21.4    & \cellcolor[HTML]{FDF0EF}24.7     & \cellcolor[HTML]{FCEEED}26      & \cellcolor[HTML]{EAF6F0}13.1    & \cellcolor[HTML]{F9FCFB}14.4    & \cellcolor[HTML]{FFFFFF}14.9    & \cellcolor[HTML]{FEF8F7}19.9    & \cellcolor[HTML]{FDF2F1}23.7    & \cellcolor[HTML]{B2DFC9}8.1     \\
\microsoftstt                      & \multicolumn{1}{l}{-}           & \cellcolor[HTML]{FCEFEE}25.4    & \cellcolor[HTML]{F9DCD9}38.1     & \multicolumn{1}{l}{-}           & \cellcolor[HTML]{D7EFE3}11.4    & \cellcolor[HTML]{FAE1DF}34.5     & \multicolumn{1}{l}{-}           & \multicolumn{1}{l}{-}            & \cellcolor[HTML]{F8D7D5}40.9    & \cellcolor[HTML]{FBE5E4}31.9    & \multicolumn{1}{l}{-}           & \multicolumn{1}{l}{-}           & \cellcolor[HTML]{FCEBEA}28      & \multicolumn{1}{l}{-}           & \cellcolor[HTML]{FCF0EF}25.2    \\
\omniaasroneb                      & \cellcolor[HTML]{FBE9E8}29.2    & \cellcolor[HTML]{FBE9E7}29.7    & \cellcolor[HTML]{FAE4E2}32.8     & \cellcolor[HTML]{F8DBD8}38.9    & \cellcolor[HTML]{FFFFFF}14.9    & \cellcolor[HTML]{FCECEB}27.3     & \cellcolor[HTML]{F6D0CD}45.7    & \cellcolor[HTML]{F6CDCA}47.7     & \cellcolor[HTML]{F3BDB8}58.4    & \cellcolor[HTML]{FBE6E5}31.2    & \cellcolor[HTML]{E98C84}89.8    & \cellcolor[HTML]{F9DEDC}36.5    & \cellcolor[HTML]{F5CBC7}49      & \cellcolor[HTML]{F3BEBA}57.3    & \cellcolor[HTML]{FFFCFC}17.2    \\
\omniaasrsevenb                      & \cellcolor[HTML]{FCEFEE}25.3    & \cellcolor[HTML]{FDF3F3}22.8    & \cellcolor[HTML]{FBE6E4}31.4     & \cellcolor[HTML]{FAE2E0}34.1    & \cellcolor[HTML]{F1F9F5}13.7    & \cellcolor[HTML]{FCEEED}26.3     & \cellcolor[HTML]{F8DAD8}39.2    & \cellcolor[HTML]{F6CCC9}48.2     & \cellcolor[HTML]{F5C6C2}52      & \cellcolor[HTML]{FCEEED}26.3    & \cellcolor[HTML]{EFA7A1}72.6    & \cellcolor[HTML]{FAE3E1}33.3    & \cellcolor[HTML]{F7D4D1}43.1    & \cellcolor[HTML]{F5C8C5}50.7    & \cellcolor[HTML]{FFFEFE}16      \\
\sarvamaudio                       & \cellcolor[HTML]{E6F4ED}12.7    & \cellcolor[HTML]{9BD6B9}6.1     & \cellcolor[HTML]{FEF6F6}20.9     & \cellcolor[HTML]{E7F5EE}12.8    & \cellcolor[HTML]{8FD1B1}5       & \cellcolor[HTML]{FFFBFB}17.6     & \cellcolor[HTML]{FFFDFD}16.3    & \cellcolor[HTML]{FDF0EF}24.8     & \cellcolor[HTML]{FEF9F9}18.9    & \cellcolor[HTML]{C0E5D3}9.4     & \cellcolor[HTML]{F4FAF7}14      & \cellcolor[HTML]{D5EEE1}11.2    & \cellcolor[HTML]{F7FBF9}14.2    & \cellcolor[HTML]{FFFAFA}18.2    & \cellcolor[HTML]{A5DAC0}7       \\
\sarvamsaarika                        & \multicolumn{1}{l}{-}           & \cellcolor[HTML]{B3E0CA}8.2     & \cellcolor[HTML]{FBE9E7}29.6     & \cellcolor[HTML]{F4FAF7}14      & \cellcolor[HTML]{9CD7BA}6.2     & \cellcolor[HTML]{FCEEED}26       & \cellcolor[HTML]{FFFDFD}16.4    & \multicolumn{1}{l}{-}            & \cellcolor[HTML]{FEF9F9}18.9    & \cellcolor[HTML]{C7E8D8}10      & \cellcolor[HTML]{FFFFFF}15.1    & \cellcolor[HTML]{E3F4EC}12.5    & \cellcolor[HTML]{FFFFFF}14.9    & \cellcolor[HTML]{FEF9F9}18.9    & \multicolumn{1}{l}{-}          \\
\bottomrule
\end{tabular}

\label{tab:main-table}


\end{subfigure}
\hfill
\begin{subfigure}[c]{0.25\textwidth}
\centering
\includegraphics[width=\linewidth]{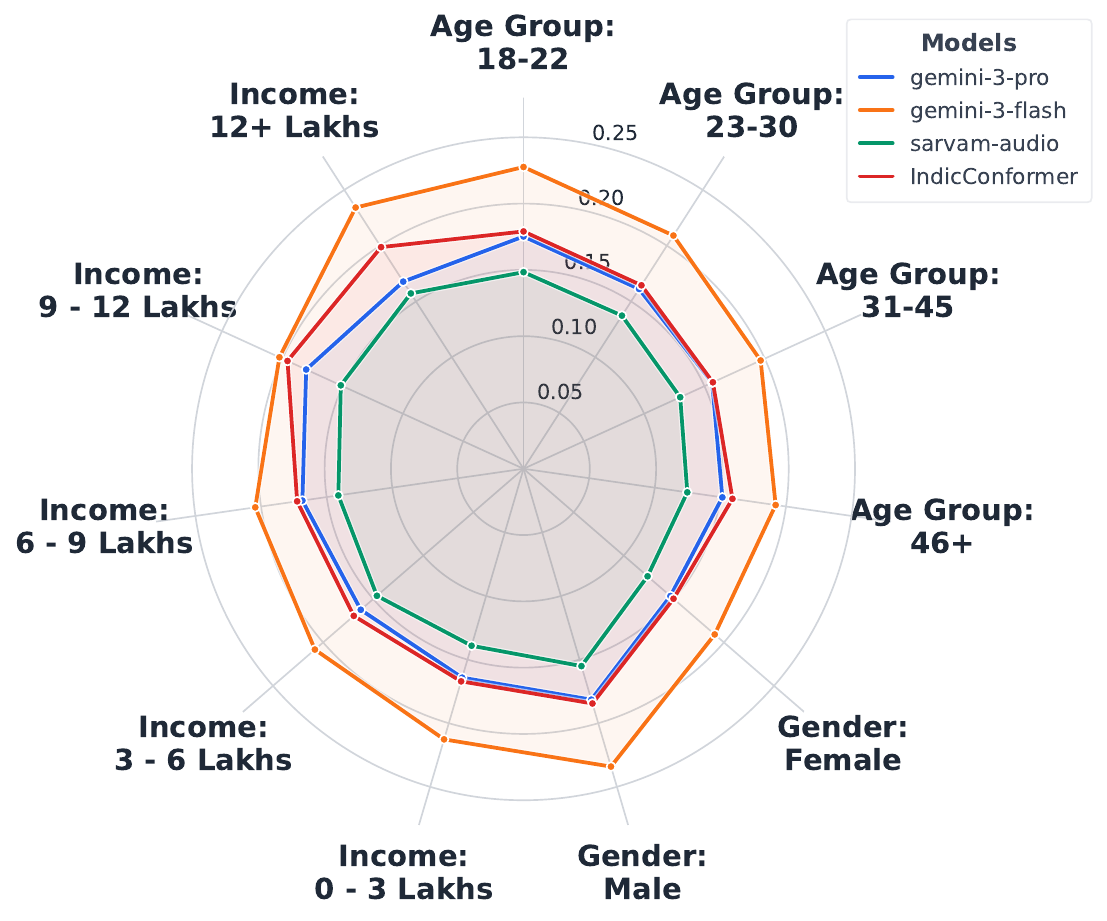}

\caption{Averaged Language-wise WER across models by age, gender, and income.}
\label{fig1}
\end{subfigure}

\end{figure*}

\section{The Voice of India Benchmark}

\subsection{Speech Data Collection}

\noindent\textbf{Platform and Contributor Onboarding.}
Speech data was collected through an online platform enabling large scale remote participation, where contributors across India recorded dual-channel, two-person conversations on assigned everyday-life topics through a peer to peer interface. Recruitment was conducted through the Josh Talks community, a large nationwide digital platform with millions of users distributed across the country, enabling outreach to speakers from diverse geographic regions, including rural and semi urban areas that are typically underrepresented in speech datasets. The final dataset contains contributions from over 36,000 speakers across 15 languages.

Collecting speech from such a geographically dispersed population introduced significant technical and logistical challenges. Contributors often used low end smartphones and unstable internet connections, particularly in low bandwidth rural environments, requiring the recording infrastructure to operate reliably under such conditions. To ensure recording quality, contributors first completed a screening task assessing language familiarity before being granted access to recording tasks. Approved participants were compensated for their contributions. All participants provided informed consent, and the data collection protocol was approved by the institute’s internal ethics committee.

\noindent\textbf{Topic Design and Prompt Generation.}
Eliciting spontaneous speech at scale is challenging, as contributors often produce short responses without structured guidance. To encourage natural, extended speech with diverse vocabulary, we curated a repository of conversational prompts spanning domains such as everyday life, personal experiences, travel, education, and social interactions. Each topic seeds a two-person conversation, with an open ended narrative cue and progressively revealed follow up questions that loosely guide the exchange toward richer descriptions and reflections rather than scripting it.

To generate prompts across 15 languages, we used GPT 4.5 to produce candidate topics across domains such as finance, healthcare, agriculture, and digital services. All machine generated prompts were reviewed and refined by language experts to ensure linguistic naturalness and cultural relevance. The final repository contains over 1,000 topics per language, translated and localized to preserve cross language comparability while retaining region specific characteristics.

\noindent\textbf{Audio Segmentation and Quality Control.}
Raw recordings were segmented into utterances using WebRTC VAD, with adjacent speech regions merged based on short silences and duration limits. Segments that were too short or excessively long were discarded. Automated language identification using Meta MMS~\cite{pratap2024scaling} and SpeechBrain VoxLingua107~\cite{speechbrain} filtered mislabeled audio, yielding approximately 1,000 hours per language. Finally, acoustic quality was enforced using DNSMOS~\cite{reddy2021dnsmos}, removing segments with low perceptual quality scores.


\noindent\textbf{Demographically Stratified Sampling.}
To ensure proportional representation, we used cluster-based sampling where districts were grouped into geographically and dialectally coherent clusters. Data volume per cluster was aligned with population proportions from the 2011 Census of India~\cite{chandramouli2011census}. Within each demographic quota, segments were prioritized using inverse-frequency word weights i.e., rare words (1-100 occurrences) received the highest weight (50), while mid-frequency (101-300) and moderately frequent words (301-1000) were assigned weights of 20 and 5 respectively, and common words ($>$1000) receiving a weight of 0.5. Segment scores were computed as the mean weight of their constituent words, allowing the selection process to favor segments with richer and more diverse vocabulary while maintaining demographic balance.



\subsection{Transcription Process}

High fidelity reference transcripts were produced using a machine-assisted multi-annotator pipeline. Initial transcripts were generated using internally fine-tuned \textsc{Whisper} models for 11 languages and the \indicconformer \cite{mahadhwani} model for Assamese, Odia, Urdu, and Maithili. These were initially verified against the audio by a native speaker and subsequently subjected to six rounds of cross-validation by independent annotators. Any segments flagged as inaccurate were re-transcribed and underwent further accuracy verification, ultimately yielding highly accurate orthographic transcripts.







\subsubsection{Lattice Construction}

Following \cite{bhogale2026oiwer}, Lexical and phonetic variations were generated and validated through a dedicated pipeline operating in parallel with human transcription.

\textbf{Variation generation.} We prompted Gemini 3 Flash to exhaustively enumerate valid word substitutions given both the ground-truth transcript and all model hypotheses. Segmentation variants (e.g., \textit{login} vs.\ \textit{log in}) and named-entity forms were captured by instructing the model to treat multi-word phrases as atomic units.

\textbf{Pruning.} To remove invalid variations obtained from the first stage, we re-prompted Gemini 3 Flash to re-evaluate the generated lattice, by prompting the model to strictly align with the original ground-truth semantics.

\textbf{Consensus alignment.} For contiguous error spans, cases where at least four of the eight top-performing models agreed on a hypothesis absent from the lattice were flagged. Flagged spans with semantic similarity below 0.5 (computed via fine-tuned BERT) \cite{deode2023l3cube} were submitted for human review; acoustically ambiguous segments confirmed by annotators were incorporated into the lattice unconditionally.

\textbf{Disfluency handling.} Verbatim human transcripts retained half-words and disfluencies. To avoid penalizing intelligibility-oriented models, such elements were made optional by merging adjacent lattice nodes. Conversely, low-amplitude sounds consistently transcribed by at least four models but absent from human references were added as optional nodes.


\section{Experiment Setup}
\subsection{Models Evaluated}

We evaluate 14 ASR systems, including 11 proprietary APIs and 3 open source models. A model is evaluated for a language only if it provides explicit support through a native language tag or can be reliably conditioned through prompts indicating the target language. For dialects such as Bhojpuri and Chhattisgarhi, dialect specific tags are used when available; otherwise prompt based conditioning is used, with Hindi as the fallback following the 2011 Census of India classification.

The evaluated systems include proprietary APIs from Sarvam (\sarvamaudio, \sarvamsaarika), Google (\geminipro, \geminiflash), OpenAI (\gptfouro, \gptfouromini), \amazontranscribe, \deepgramnova, \assemblyai, \elevenlabscribe, and \microsoftstt. We also include three open source models: \indicconformer~(AI4Bharat) and Meta's \omniaasroneb~ and \omniaasrsevenb. All models are evaluated using their default inference configurations.

\subsection{Evaluation Metric}
We evaluate models using Orthographically-Informed Word Error Rate (OIWER)~\cite{bhogale2026oiwer} metric. Unlike standard WER, this accounts for permissible spelling variation between hypothesis and reference which reduces spurious errors caused by orthographic variation and better reflects recognition quality in languages with flexible spelling conventions.

\section{Results and Discussion}

\subsection{Evaluation of models on the Voice of India Benchmark}

Table~\ref{tab:main-table} shows that most models exceed a WER of 20 (highlighted in red), a threshold often associated with practical usability, and no system meets this criterion consistently across all languages. Even the best performing model, \sarvamaudio, exceeds this threshold on Bhojpuri ($20.9$) and Maithili ($24.8$). \sarvamaudio~achieves the lowest WER in 13 of 15 languages, followed by \sarvamsaarika~and \geminipro. \indicconformer~and \elevenlabscribe~show moderate performance, while several models perform substantially worse; in some cases \assemblyai~exceeds WER 100, indicating transcription failure. Notable anomalies also appear: \geminipro~performs best on dialectal varieties such as Bhojpuri and Chhattisgarhi, while \gptfouromini~shows severe degradation on Gujarati ($295.9$) and Malayalam ($167.8$) despite moderate performance on Hindi. These results highlight the  difficulty of robust ASR across Indian languages and dialects, reflecting their limited representation in large scale multilingual training and the challenges introduced by script diversity and orthographic flexibility.

\subsection{Does the performance vary across regions of India?}

District level WER shows strong geographic variation, ranging from about 4\% (Nainital) to 44\% (Mannarakkat). Higher error rates appear in parts of South India, particularly Kerala and interior Karnataka, and in North Bihar, reflecting the presence of underrepresented languages such as Maithili and Bhojpuri. In contrast, districts across the Hindi belt (Uttar Pradesh, Delhi, Haryana, Rajasthan, and Madhya Pradesh) cluster below 10\% WER, indicating stronger alignment with standard Hindi speech. Metropolitan districts also tend to show lower error rates. Overall, the pattern reveals a clear geographic bias, with linguistically diverse or underrepresented regions exhibiting substantially higher WER than the Hindi belt and major urban centers.

\subsection{Are existing Indic ASR benchmarks reliable?}

\begin{figure}[h!]
    \centering
    \includegraphics[width=0.8\linewidth]{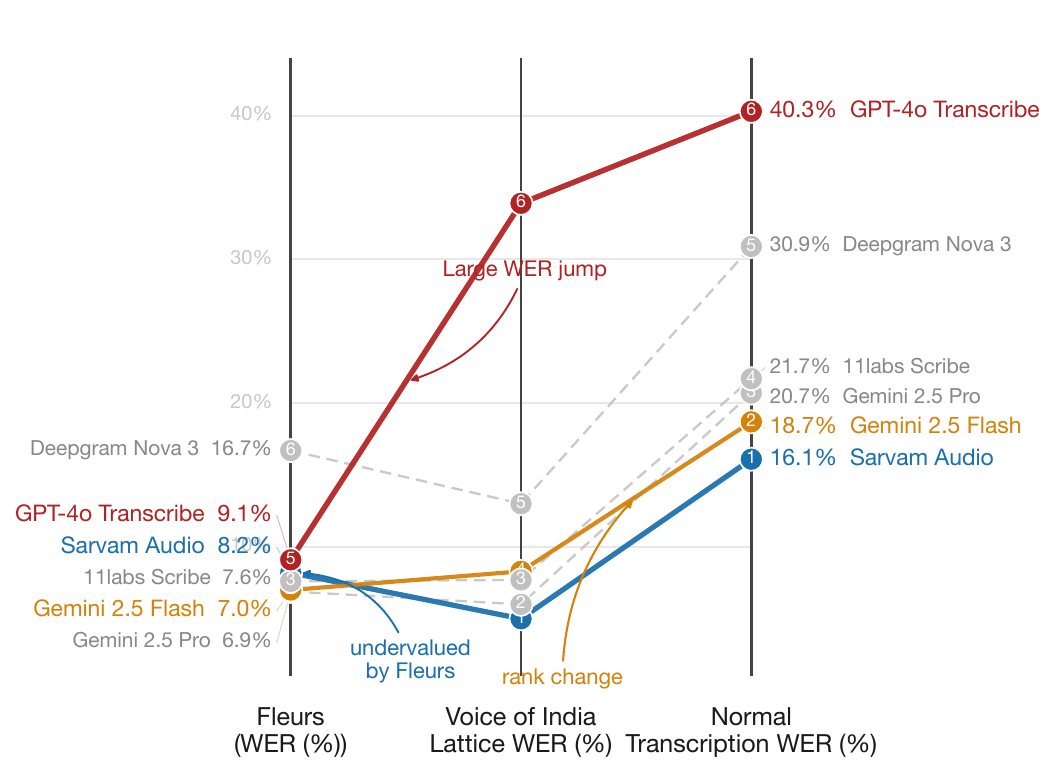}
    \caption{WER of six models across FLEURS (public benchmark), VoI Lattice, and Normal Transcription (single-reference), with circled rank badges (1 = lowest WER). }
    \label{gamify}
\end{figure}

Public benchmarks such as FLEURS \cite{conneau2022}, though widely used for reporting performance, are vulnerable to overfitting due to their static and publicly accessible evaluation sets. As shown in Figure \ref{gamify}, models that achieve strong WER on FLEURS often perform substantially worse on our benchmark, particularly for morphologically richer languages. Moreover, single reference WER is sensitive to transcription style, penalizing orthographic differences despite correct acoustic modeling. The lattice based evaluation mitigates this issue by allowing multiple valid spelling variants, producing more stable system rankings. 

\subsection{Does WER vary across different audio attributes?}

\begin{figure}[t]
    \centering
    \includegraphics[width=0.8\linewidth]{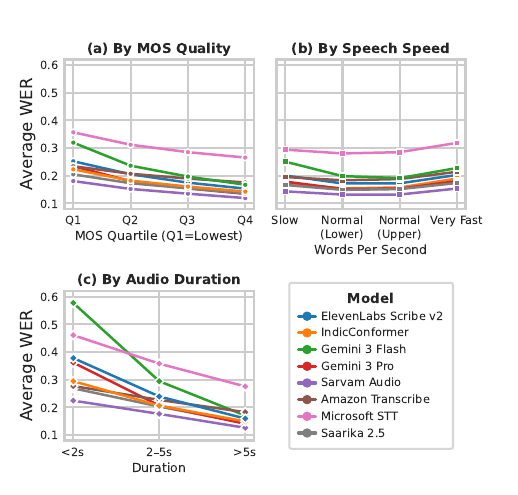}
    \caption{Performance of transcription models (cross-lingual average WER) across speech characteristics, including MOS quality quartiles, speech rate, and audio duration.}
    \label{fig2}
\end{figure}

Figure \ref{fig2} shows that deviations from ideal acoustic conditions consistently increase error rates across DNSMOS \cite{reddy2021dnsmos} quality quartiles, speaking-rate quartiles, and utterance duration bins ($<$2s, 2--5s, $>$5s). Audio degradation raises WER monotonically; ElevenLabs Scribe rises from $15.31\%$ to $25.20\%$ and Gemini-3-Pro from $13.42\%$ to $23.44\%$ between the highest and lowest quality quartiles. Speaking rate exhibits a U-shaped pattern, with Indic Conformer WER peaking at $27.57\%$ (slow) and $27.53\%$ (very fast) versus $24.75\%$ at moderate speeds. Short utterances are most affected due to limited semantic context; Amazon STT degrades from $10.45\%$ ($>$5s) to $18.74\%$ ($<$2s), with Microsoft STT showing a similar trend ($10.90\%$ to $18.59\%$).

\subsection{Are models fair across demographics?}

A fair model achieves comparable average WER across demographic groups (Figure \ref{fig1}). While none of the models exhibit strict parity, the observed differences are small, indicating broadly consistent performance. Models perform slightly better on female speech (3.1 to 4.3 percent gap), while younger speakers (18 to 22) show higher error rates than older speakers (46 plus). Income based differences are also minor, with slightly higher WER for higher income speakers, possibly due to increased linguistic complexity such as code mixing. Notably, \geminipro~and \sarvamaudio~show less than 2 percent variance across income groups.



\subsection{Recommendations for Model Developers}

Based on our analysis, we group evaluated systems into three tiers according to the nature of their failures, and provide targeted recommendations for each category.

\noindent\textbf{Tier I: Top-performing systems (Ranks 1--6, WER $\leq$ 20\%).}
Tier I models (Sarvam-Audio, Gemini-3-Pro, IndicConformer) largely solve general multilingual transcription but reveal three critical fault lines: (1) low-resource languages like Bhojpuri and Maithili suffer WERs 4-5$\times$ higher than Hindi; (2) a systematic 3.1--4.3\% male-speaker penalty persists across all architectures; and (3) models fail catastrophically for out-of-region migrants (e.g., Chattisgarhi speakers in Tamil Nadu face WERs of 55\%-65\%). Resolving these gaps demands targeted regional data collection, gender-stratified training, and explicit cross-regional evaluation metrics.

\noindent\textbf{Tier II: Competent but fragile systems (Ranks 7--8, WER 21\%--30\%).}
These systems perform well on canonical speech but degrade sharply under distributional shifts. For example, Gemini 3 Flash exhibits a severe short-utterance penalty compared to longer segments, as well as significant performance degradation on low-quality audio, necessitating dedicated short-audio pathways and multi-condition training. Concurrently, Microsoft STT struggles with underrepresented languages, demonstrating noticeably higher error rates on languages such as Bhojpuri and Malayalam. Improving overall robustness requires targeted SNR augmentation and expanded data coverage, particularly for the Eastern Hindi belt and Dravidian language families.

\noindent\textbf{Tier III: Inadequate systems (Ranks 9--16, WER $\geq$ 35\%).}
These models face severe coverage limitations, especially for Dravidian languages. Deepgram Nova-3 and OmniASR yield elevated error rates in Tamil (WER: 67.8\%) and Odia (WER: 89.8\%), while GPT-4o-mini and AssemblyAI Universal produce extreme errors in Gujarati (WER: 297\%) and Malayalam/Telugu (WER $\ge$ 100\%), driven by failed language detection and hallucinations rather than standard transcription mistakes. 
Consequently, substantial retraining and targeted language adaptations are essential prior to broad deployment.

\section{Conclusion}

We introduce Voice of India, a benchmark for evaluating ASR systems on real world Indian speech collected from unscripted telephonic conversations across multiple languages and regions. The benchmark incorporates multiple transcription variants and evaluates systems using orthographically informed WER to better reflect natural spelling variation in spontaneous speech. Evaluation across state of the art systems reveals substantial robustness gaps, with large performance differences across languages and regions and particularly high error rates in linguistically diverse areas. We further show that public benchmarks can overestimate real world performance and that single reference WER exaggerates errors caused by orthographic variation. 

\section{Generative AI Use Disclosure}
Generative AI tools were used solely for language polishing and editing during the preparation of this manuscript. These tools assisted with improving clarity, grammar, and conciseness of the writing. No generative AI system was used to generate experimental results, analyses, figures, or scientific conclusions. All technical content, experiments, and interpretations were developed and verified by the authors.

\section{Acknowledgments}

We thank the contributors and annotators whose efforts made the Voice of India benchmark possible.


\bibliographystyle{IEEEtran}
\bibliography{references}

\end{document}